\pdfoutput=1

\documentclass[11pt]{article}

\usepackage{latex/acl}

\usepackage{times}
\usepackage{latexsym}

\usepackage{amsmath}
\usepackage{amsfonts}
\usepackage{booktabs}
\usepackage{multirow}
\usepackage{amssymb}
\usepackage{lipsum}
\usepackage{duckuments}
\usepackage{pifont}
\newcommand{\xmark}{\ding{55}}%

\usepackage[T1]{fontenc}

\usepackage[utf8]{inputenc}

\usepackage{microtype}

\usepackage{inconsolata}
\usepackage{graphicx}
\title{Extending CLIP's Image-Text Alignment to Referring Image Segmentation}

\author{
   Seoyeon Kim$^1$\,\,
   Minguk Kang$^1$\,\,
   Dongwon Kim$^1$ \,\,
   Jaesik Park$^2$\,\,
   Suha Kwak$^1$\,\,\vspace{1mm}\\
    $^1$POSTECH,~~
    $^2$Seoul National University \\
    $^1$\normalsize{\texttt{\{syeonkim07,mgkang,kdwon,suha.kwak\}@postech.ac.kr}} \\
    $^2$\normalsize{\texttt{jaesik.park@snu.ac.kr}}
}

\begin{document}
\maketitle
\begin{abstract}
Referring Image Segmentation (RIS) is a cross-modal task that aims to segment an instance described by a natural language expression. Recent methods leverage large-scale pretrained unimodal models as backbones along with fusion techniques for joint reasoning across modalities. However, the inherent cross-modal nature of RIS raises questions about the effectiveness of unimodal backbones. We propose RISCLIP, a novel framework that effectively leverages the cross-modal nature of CLIP for RIS. Observing CLIP's inherent alignment between image and text features, we capitalize on this starting point and introduce simple but strong modules that enhance unimodal feature extraction and leverage rich alignment knowledge in CLIP's image-text shared-embedding space. RISCLIP exhibits outstanding results on all three major RIS benchmarks and also outperforms previous CLIP-based methods, demonstrating the efficacy of our strategy in extending CLIP's image-text alignment to RIS.
\vspace{+2.5mm}
\end{abstract}
\section{Introduction}
 
Referring Image Segmentation (RIS) is a multi-modal task that aims to produce a pixel-wise mask of an instance referred to by a natural language expression. The task holds great potential with various applications, such as language-based image editing~\citep{chen2018languagebased, parmar2023zero, brooks2022instructpix2pix} and human-robot interaction~\citep{wang2019reinforced}.

\begin{figure}[t!]
    \centering
    \includegraphics[width=0.97\linewidth]{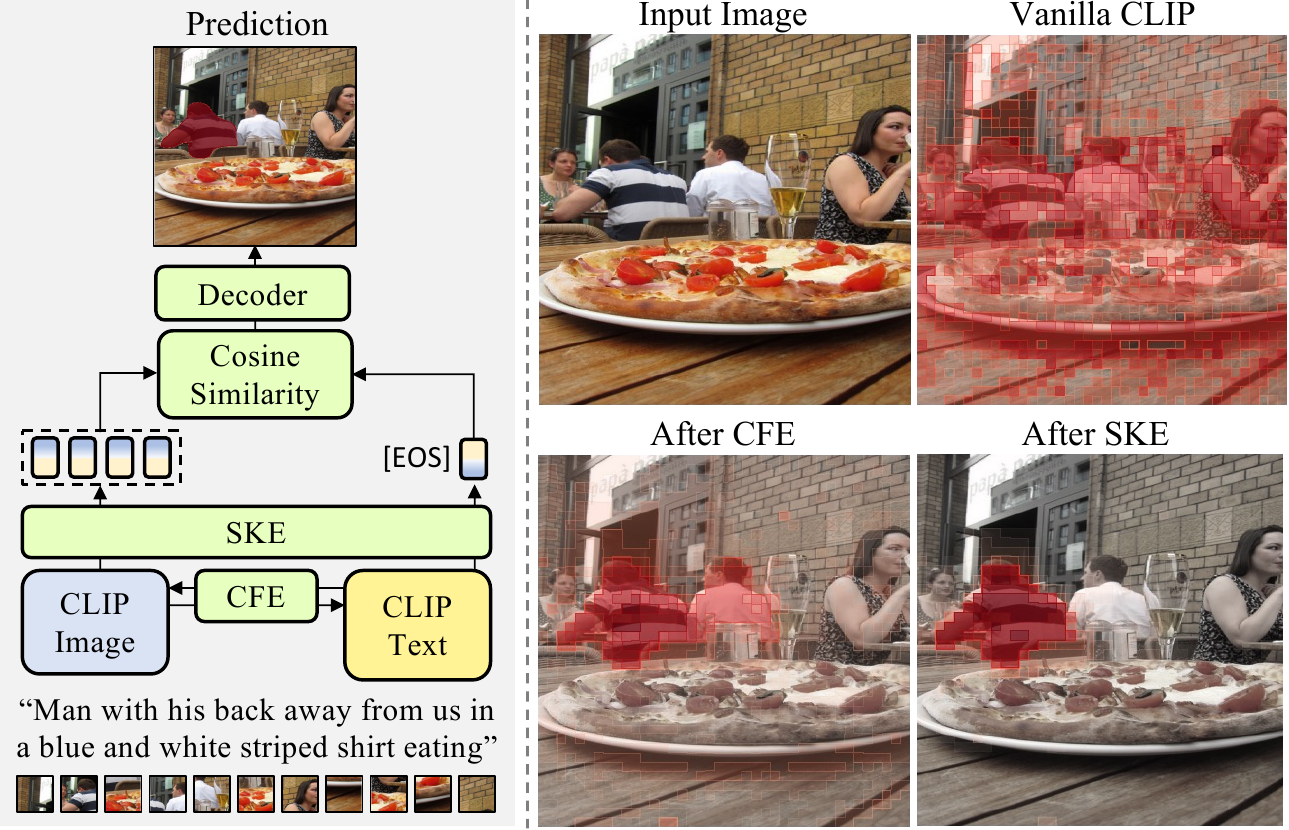}
    \caption{CLIP's image-text alignment produces preliminary patch-level groundings through cosine similarity between patch-level image and sentence-level text features. Building upon this alignment, we refine CLIP's groundings into accurate segmentations with three modules. Cross-modal Feature Extraction (CFE) modules enhance CLIP's unimodal image and text features by aligning them at candidate regions. Shared-space Knowledge Exploitation (SKE) modules leverage the rich alignment knowledge in CLIP's image-text shared-embedding space to discern the target referent. Lastly, a decoder transforms the patch-level grounding into a pixel-wise segmentation.}
    \label{fig:teaser}
\end{figure}
RIS poses a formidable challenge, demanding a nuanced understanding of both visual and linguistic modalities. Thus, conventional methods~\citep{li2021referring, wang2022cris, zhu2022seqtr, yang2022lavt, liu2023polyformer} leverage the profound knowledge learned by large-scale pretrained models, employing image and text encoders as backbones, such as Swin-T \citep{liu2021swin} trained on ImageNet-21K \citep{ridnik2021imagenetk} and BERT \citep{devlin2018bert} trained on Wikipedia and Google’s BookCorpus. Additionally, various fusion techniques \citep{hu2016segmentation, ye2019cross, ding2021vision, hu2020bi, hui2020linguistic} have been introduced to enable joint reasoning across modalities, significantly advancing RIS.

However, the inherent cross-modal nature of RIS raises questions about the effectiveness of unimodal backbones. In contrast, we posit that the cross-modal nature of CLIP~\citep{radford2021learning} makes it a better candidate for RIS. Extending on the findings of MaskCLIP~\cite{zhou2022extract}, we observe that CLIP possesses image-text alignment beneficial for RIS---cosine similarity maps between patch- and sentence-level features~\citep{zhou2022extract} produce preliminary groundings. To fully utilize this image-text alignment, we freeze CLIP and build upon this promising starting point. While previous work~\citep{wang2022cris, xu2023bridging} demonstrates the potential of adopting CLIP for RIS, their performances trail behind the current state of the arts, indicating an opportunity for improvement.

In our framework, \textit{RISCLIP}, we effectively adapt CLIP to RIS, capitalizing on its image-text alignment. Firstly, we enhance the unimodal feature extraction by introducing cross-modal interaction between the image and text encoders with Cross-modal Feature Extraction (CFE) modules. These modules effectively align the image and text features at candidate regions---regions described by or related to the target text. Then, we leverage the rich alignment knowledge captured in CLIP's image-text shared-embedding space by introducing inter- and intra-modal interactions after the feature extraction process with Shared-space Knowledge Exploitation (SKE) modules. The comprehensive interactions allow RISCLIP to discern the target from the candidate regions. Our CFE and SKE modules effectively adapt CLIP to RIS, evolving CLIP's preliminary image-text alignment into accurate groundings, as shown in Fig.~\ref{fig:teaser}.

RISCLIP exhibits outstanding performance on all three major RIS benchmarks. Particularly, RISCLIP excels on the more challenging datasets, such as RefCOCOg~\cite{mao2016generation} with complicated texts. Such result indicates that adopting a cross-modal backbone like CLIP, which trains on varied captions including extensive expressions, is beneficial for RIS. Furthermore, RISCLIP also surpasses previous CLIP-based methods~\citep{wang2022cris, xu2023bridging}, proving that such performance arises from both CLIP and our effective adaptation strategy.

\section{Related Work}
\noindent \textbf{Referring image segmentation.}
RIS aims at predicting a pixel-wise mask of an object described by a natural language text. The pioneering work~\cite{hu2016segmentation} extracts image and text features with recurrent LSTMs and a CNN and concatenates them along the channel dimension into multi-modal features. Follow-up work expands on this framework by incorporating recurrent multi-modal interactions~\cite{liu2017recurrent} along with more fine-grained segmentation with hierarchical visual features~\cite{li2018referring, margffoy2018dynamic, chen2019see, Jain_2022}. Another line of research focuses on attending to more important words in the referring expression~\cite{yu2018mattnet, shi2018key, liu2021cross} and proposes effective cross-modal attention modules~\cite{ye2019cross, ding2021vision, hu2020bi, hui2020linguistic}. Recent methods adopt pretrained transformer encoders to extract image and text features~\cite{kim2022restr, tang2023contrastive}, and others further leverage the encoder transformer layers for multi-modal feature extraction~\cite{feng2021encoder, yang2022lavt, zhang2022coupalign, OuYang2023SLViTSL}. Moving towards real-world conditions, recent work tackles settings where expressions describe none to multiple objects~\cite{Hu_2023_ICCV, liu2023gres}, image-expression pairs only are provided without segmentation masks~\cite{strudel2022weaklysupervised, liu2023gres, kim2023shatter, lee2023weakly}, and no labels are provided for training~\cite{yu2023zeroshot, suo2023text}.

\noindent \textbf{Contrastive language-image pre-training.} CLIP~\cite{radford2021learning} is well-known for its general cross-modal capacity. Acquired through extensive contrastive pretraining on large-scale image-text pairs, CLIP carries not only expertise knowledge in both visual and linguistic modalities but also general alignment between image and text features. Various multi-modal tasks, including text-to-image generation~\cite{ramesh2022hierarchical, rombach2022high} and visual captioning~\cite{mokady2021clipcap, hessel2021clipscore}, benefit from CLIP's rich multi-modal alignment. Several works attempt to adapt CLIP to dense prediction tasks, such as open vocabulary object detection~\cite{du2022learning, Hanoona2022Bridging} and semantic segmentation~\cite{Luo2022SegCLIP, xu2023side}. In particular, MaskCLIP~\cite{zhou2022extract} exploits the alignment between patch- and sentence-level features for zero-shot open vocabulary segmentation. We hypothesize that such patch-level alignment is a good starting point for RIS, and, consistent with such hypothesis, observe that the alignment produces a noticeable 23.86 mIoU on the most challenging RIS benchmark---RefCOCOg-UMD~\cite{nagaraja2016modeling}---with a upsampling decoder attached. Thus, we propose a new framework that effectively exploits such informative cross-modal alignment of CLIP to produce accurate RIS predictions.

\noindent \textbf{CLIP for RIS.}
Although methods that adopt CLIP for RIS exist~\cite{radford2021learning}, they either fully finetune CLIP and risk losing its general knowledge~\cite{wang2022cris} or do not explicitly leverage the alignment between the image and text features learned from millions of image-text pairs~\cite{xu2023bridging}. Above all, their performance falls behind the state of the art, suggesting room for improvement. Thus, we take a new approach of explicitly exploiting CLIP's rich image-text alignment by extracting preliminary grounding maps from frozen CLIP and enhancing them into accurate segmentations with our adaptive modules. Our framework achieves compelling results, showing that we effectively extend CLIP's image-text alignment to RIS.
\section{Method}
\label{sec:ACGAN}

\begin{figure*}[t!]
    \centering
    \includegraphics[width=0.97\linewidth]{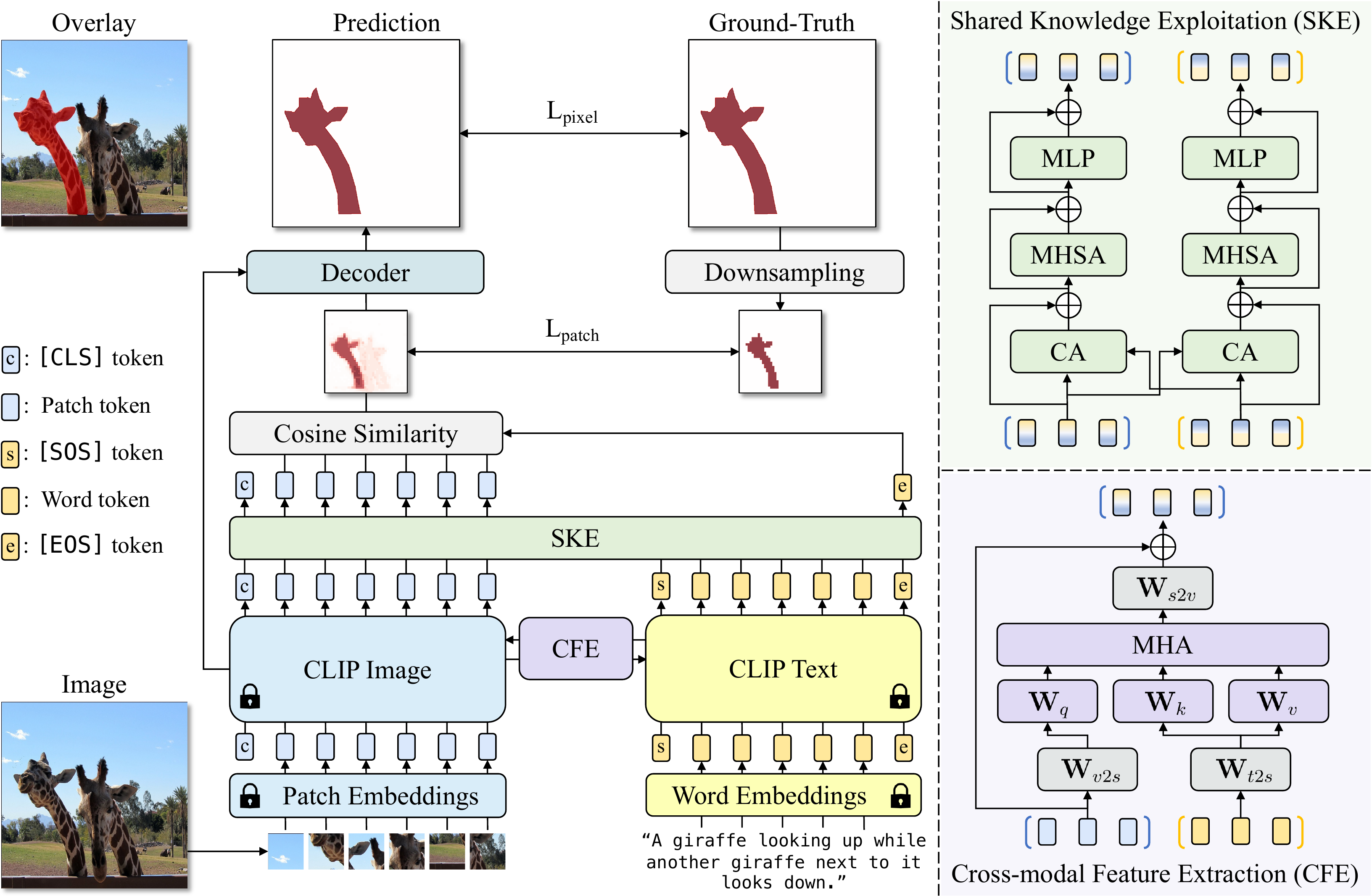}
    \caption{The overall pipeline of RISCLIP. We adopt frozen CLIP image and text encoders as backbones to exploit their aligned image and text features and adapt them to RIS with two modules, CFE and SKE. Firstly, the CFE modules between the encoders enable cross-modal commnuication between the two encoders to align their unimodal features at candidate regions. Secondly, the SKE modules on top of the encoders leverage the rich cross-modal alignment knowledge in CLIP's image-text shared embedding space to discern the target referent. Then a cosine similarity between the patch- and sentence-level features produces a patch-level grounding map. Lastly, a decoder refines the map into a pixel-level segmentation prediction.}
    \label{fig:overall}
\end{figure*}
\label{sec:Overview}
Fig.~\ref{fig:overall} illustrates the overall pipeline of our method, RISCLIP. We exploit CLIP's cross-modal alignment between patch- and sentence-level features and evolve cosine similarity maps between them into precise pixel-wise groundings with our new framework. The Cross-modal Feature Extraction (CFE) modules enhance the unimodal feature extraction of CLIP with cross-modal communication, aligning image and text features at candidate regions related to the text. The Shared-space Knowledge Exploitation (SKE) modules exploit CLIP's rich knowledge captured within the image-text shared embedding space to discern the target referent from candidate regions, particularly those described by complicated expressions. Together, CFE and SKE modules adapt CLIP to RIS, producing precise patch-level grounding maps. Finally, a simple decoder refines these maps into pixel-level segmentations. The following sections detail each module—CFE, SKE, and the decoder—starting with the original CLIP feature extraction.

\subsection{CLIP for referring image segmentation}
\label{sec:clip}
To fully utilize CLIP's invaluable image-text alignment, we freeze CLIP and introduce modules that enhance the features for RIS. We explain the feature extraction process of CLIP and detail the construction of the patch-level grounding map in the following paragraphs. 

\noindent \textbf{Feature extraction.}
Both CLIP's image and text encoders consist of repeated transformer layers~\cite{vaswani2017attention}, followed by a layer normalization (LN)~\cite{ba2016layer} and a linear projection to a shared image-text embedding space. Each transformer layer has two submodules: a multi-head self-attention (MHSA) and a multilayer perceptron (MLP) with each submodule preceded by LN. Text features are extracted via a text encoder as a sequence of word- and sentence-level representations. Firstly, the sentence is transformed into word embeddings via byte pair encoding~\cite{gage1994new} and encased with a learnable [\texttt{SOS}] and [\texttt{EOS]} token. This sequence is then passed through transformer layers and linearly projected to the shared image-text embedding space. The output [\texttt{EOS}] token acts as the sentence-level representation of the text, and we denote it $\mathbf{{t}_{\text{eos}}} \in\mathbb{R}^{1\times d}$, where $d$ is the dimensionality of the shared embedding space. Analogously, image features are extracted with an image encoder as a sequence of patch-level representations prepended with an image-level embedding. Specifically, the image is divided into a sequence of patches, prepended with a learnable [\texttt{CLS]} token, passed through transformer layers, and linearly projected to the shared embedding space. We denote the output patch-level features as $\mathbf{V}_{\text{patch}}\in\mathbb{R}^{N_{\text{visual}}\times d}$, where $N_{\text{visual}}$ is the number of patches. 

\noindent \textbf{Patch-level grounding map.}
We build upon the findings of  MaskCLIP~\cite{zhou2022extract} that CLIP possesses cross-modal alignment between patch- and sentence-level features to produce preliminary patch-level RIS groundings. Slightly modifying the image feature extraction process of the last image transformer layer, we adopt the value tokens from the MHSA as patch tokens and pass them through the subsequent LN, MLP, and linear projection to the image-text shared embedding space to produce a new $\mathbf{V}_{\text{patch}}$. Cosine similarity between $\mathbf{V}_{\text{patch}}$ and $\mathbf{t}_{\text{eos}}$ produces preliminary grounding maps for RIS, highlighting regions related to the text. We observe that the patch-level grounding maps with a decoder attached provide a promising mIoU of 23.86 on the RefCOCOg-UMD test set~\cite{nagaraja2016modeling}. Adopting this as a good starting point, we introduce modules to transform the preliminary maps into accurate segmentations. 

\noindent \textbf{Adapters.}
Since the CLIP backbone is trained with contrastive learning between the image- and sentence-level features, its features are suboptimal for dense prediction tasks like RIS. Thus, we introduce adapters to transform the features into representations more appropriate for RIS. We adopt the adapter architecture from~\cite{pmlr-v97-houlsby19a, chen2022adaptformer}, which consists of a down-projection linear layer, a non-linear activation, and an up-projection linear layer. These structures are residually attached after the MHSA and MLP modules of CLIP's transformer layers. The residual summations introduce fine-grained structural details that refine the segmentation boundaries of the patch-level grounding map, thereby increasing the mIoU on the RefCOCOg-UMD test set to 48.29 mIoU.  

\subsection{Cross-modal feature extraction}
\label{sec:cmfe}
The independent feature extraction of image and text features in CLIP yields unimodal features which are bound to limited alignment between the target instance and referring text. Consider the image-text pair in Fig.~\ref{fig:overall}. Given an image of two giraffes and the text ``A giraffe looking up while another giraffe next to it looks down'', the patch-level features of the target giraffe are unlikely to perfectly align with the sentence-level feature, as the giraffe can be described with different texts like ``the taller giraffe behind the fence'' and ``giraffe sticking its chin up.'' For the target patch-level features to better align with the sentence-level feature, they must evolve to be like the text feature, or vice versa, through cross-modal interaction. Therefore, we introduce Cross-modal Feature Extraction (CFE) modules between the unimodal image and text encoders to enable cross-modal communication via cross-attention which aligns the image and text features at text-relevant regions. Starting from the deepest layers of the backbone, we pair an image and text encoder layer and attach a single CFE module in between to communicate intermediate image and text features, where the number of CFE modules introduced is a hyperparameter.

Consider a CFE module between the $k$-th text and $l$-th image transformer layers which take as input text and image features, $\mathbf{t}_{k}$ and $\mathbf{v}_{l}$, respectively. First, the CFE module projects $\mathbf{t}_{k}$ and $\mathbf{v}_{l}$ to a shared image-text embedding space with linear projections, $\mathbf{W}_{t2s}$ and $\mathbf{W}_{v2s}$, to produce $\mathbf{t}_{k}^{s}$ and $\mathbf{v}_{l}^{s}$. Then, two separate multi-head cross-attention (MHCA) modules input $\mathbf{t}_{k}^{s}$ and $\mathbf{v}_{l}^{s}$, where each modality is set as query and the other key and value, to produce text and image multi-modal features, $\mathbf{t}_{k}^{m}$ and $\mathbf{v}_{l}^{m}$. Lastly, the multi-modal features are projected from the shared embedding space back to each modalities' space with linear projections, $\mathbf{W}_{s2t}$ and $\mathbf{W}_{s2v}$ to produce the final multi-modal features $\mathbf{t}_{k}^{m'}$ and $\mathbf{v}_{l}^{m'}$. We elaborate on this process to output $\mathbf{t}_{k}^{m'}$ below, where $\mathbf{v}_{l}^{m'}$ can be computed in vice versa, as 
\vspace{-1.7mm}
\begin{align*}
    &  \mathbf{t}_{k}^{s} = \mathbf{W}_{t2s}\mathbf{t}_{k}, \quad \mathbf{v}_{l}^{s} = \mathbf{W}_{v2s}\mathbf{v}_{l}, \label{eq:eq5}\tag{1} \\
    & \mathbf{t}_{k}^{m} = \mathrm{MHCA}(\mathbf{t}_{k}^{s}, \mathbf{v}_{l}^{s}, \mathbf{v}_{l}^{s}) \label{eq:eq8}\tag{2} \\
    & \mathbf{t}_{k}^{m'} = \mathbf{W}_{s2t}\mathbf{t}_{k}^{m}.
    \label{eq:eq8}\tag{3}
\end{align*}
These multi-modal features, $\mathbf{t}_{k}^{m'}$ and $\mathbf{v}_{l}^{m'}$, are added back to the input features as $\mathbf{t}_{k} = \mathbf{t}_{k} + \mathbf{t}_{k}^{m'}$ and $\mathbf{v}_{l} = \mathbf{v}_{l} + \mathbf{v}_{l}^{m'}$ to inject multi-modal information into the CLIP features. Then, $\mathbf{t}_{k}$ and $\mathbf{v}_{l}$ are input to the $k$-th text and $l$-th image transformer layers to be processed by the subsequent MHSA and MLP. We visualize the feedforward process in the bottom-right of Fig~\ref{fig:overall}.

These modules effectively inject cross-modal information into CLIP's previously unimodal features, allowing the target patch-level features to align with the sentence-level features and vice versa. The better aligned features output more accurate groundings, increasing performance to 60.58 mIoU on the RefCOCOg-UMD test set.

\subsection{Shared-space knowledge exploitation}
Although the CFE modules effectively adapt CLIP to RIS, we observe that they often miss the target described by complex texts about intricate relationships among multiple instances. Thus, we attempt to leverage the alignment between images and lengthy captions residing in CLIP's image-text shared embedding space, which was obtained during its extensive contrastive learning pretraining. We introduce Shared-space Knowledge Exploitation (SKE) modules which execute both inter- and intra-modal conditioning on the image and text features within the shared embedding space via residually connected MHCA, MHSA, and MLPs, with LN applied before every submodule. We visualize the feedforward in the top-right of Fig.~\ref{fig:overall}. The intra-modal interactions allow the model to grasp intricate relationships between objects within in each modality, while the inter-modal interactions allow cross-modal alignment and grounding. These comprehensive interactions handle lengthy, intricate descriptions effectively, increasing the performance to 62.64 mIoU on the RefCOCOg-UMD test set.

\subsection{Decoder}
The patch-level grounding maps are upsampled to pixel-level predictions ($\mathbf{map}_{\text{pixel}}$) with a decoder. Since the role of the decoder is to figure out the boundary of the referred instance detected by the grounding map, we adopt a simple decoder~\cite{yang2022lavt}. To accurately restore the geometric details of the target referent, the decoder exploits intermediate patch-level features from the first four layers of the CLIP image backbone, $\mathbf{V}_{i}, i\in \{1, 2, 3, 4\}$. 

The decoder consists of four layers, $\mathbf{D}_{i}, i\in \{1,2,3,4\}$, where $\mathbf{D}_{i}$ comprises of two repetitions of 3$\times$3 convolutions, ReLU~\cite{agarap2018deep}, and batch normalization~\cite{pmlr-v37-ioffe15}, followed by double-resolution upsampling. The feed forward process is given by 
\begin{align*}
    & \mathbf{d}_{4} = \mathbf{D}_{4}([\mathbf{v}_{4}; \mathbf{map}_{\text{patch}}]) \\
    & \mathbf{d}_{i} = \mathbf{D}_{i}([\mathbf{d}_{i+1}; u(\mathbf{v}_{i})]), i=1,2,3, 
\end{align*}
where $\mathbf{d}_{i}$ is the output features of $\mathbf{D}_{i}$, $\mathbf{map}_{\text{patch}}$ the sigmoided patch-level grounding map, $u$ the upsampling operation, and [;] channel-wise concatenation. Finally, $\mathbf{d}_{1}$ undergoes a linear projection that produces background and foreground score maps, which are sigmoided into the pixel-wise prediction, $\mathbf{map}_{\text{pixel}}$. The binary prediction mask is obtained via argmax during inference.

\subsection{Loss functions}
RISCLIP is trained in two stages. In the first stage, the Adapters, CFE, and SKE modules align the sigmoided patch-level grounding map, $\mathbf{map}_{\text{patch}}$, with a patch-level downsampled ground truth mask, $\mathbf{mask}_{\text{patch}}$. Once $\mathbf{map}_{\text{patch}}$ converges, the decoder is introduced in the second stage to upsample $\mathbf{map}_{\text{patch}}$ to a pixel-wise map, $\mathbf{map}_{\text{pixel}}$, aligning with the pixel-level ground truth mask, $\mathbf{mask}_{\text{pixel}}$. During this stage, all modules except the decoder remain frozen, and the decoder only is trained for a single epoch.  
Although training the entire framework end-to-end with the decoder is feasible, this approach inevitably makes the decoder receive random patch-level grounding maps during initial training steps, resulting in meaningless training signals and wasted computation. Conversely, the two-stage training simplifies the decoder's role to delineating patch-level grounding boundaries, requiring only one epoch for convergence in the second stage. This approach enhances efficiency by minimizing unnecessary computation. \\
Following~\cite{li2021referring}, we adopt a linear combination of DICE/F-1 loss~\cite{milletari2016v} and focal loss~\cite{lin2017focal} for both training stages, first between $\mathbf{map}_{\text{patch}}$ and $\mathbf{mask}_{\text{patch}}$ and after between $\mathbf{map}_{\text{pixel}}$ and $\mathbf{mask}_{\text{pixel}}$.
\section{Experiments}
\subsection{Datasets and evaluation metrics}
\begin{table*}[t]  
  \caption{Comparison with state of the arts on RefCOCO~\cite{yu2016modeling}, RefCOCO+~\cite{yu2016modeling}, and RefCOCOg-UMD~\cite{mao2016generation, nagaraja2016modeling}. We reproduce DMMI~\cite{Hu_2023_ICCV} on RefCOCO and RefCOCO+ with the official code to report its mIoU scores unprovided in the original paper. RN101 is ResNet-101~\cite{He_2016_CVPR}, DN53 Darknet-53~\cite{redmon2018yolov3}, and WRN101 Wide ResNet-101~\cite{zagoruyko2016wide}. CLIP-B and CLIP-L denote the Transformer-based CLIPs which adopt ViT-B and ViT-L~\cite{vaswani2017attention} as the image encoder, respectively, while CLIP-L* is the ResNet-based CLIP which utilizes ResNet-101~\cite{He_2016_CVPR}.}
   \centering
   \resizebox{0.93\textwidth}{!}{{\begin{tabular}{lllcccccccc}
      \toprule[1pt]
      \multirow{2}*[-0.5ex]{\textbf{\normalsize{Method}}} & \multirow{2}*[0.5ex]{\textbf{Image}} &
      \multirow{2}*[0.5ex]{\textbf{Text}} & 
      \multicolumn{2}{c}{\textbf{RefCOCOg}} &  \multicolumn{3}{c}{\textbf{RefCOCO}} & \multicolumn{3}{c}{\textbf{RefCOCO+}} \\
      \cmidrule[1.0pt]( r){4-5}
      \cmidrule[1.0pt]( lr){6-8}
      \cmidrule[1.0pt]( lr){9-11}
      & \multirow{2}*[1.5ex]{\textbf{Encoder}} & \multirow{2}*[1.5ex]{\textbf{Encoder}} & Val & Test & Val & Test A & Test B & Val & Test A & Test B  \\
      \cmidrule[1.0pt]{1-11}
      \multicolumn{11}{l}{\textbf{oIoU}} \\
      \cmidrule[1.0pt]{1-11}
      BRINet~\cite{hu2020bi}    & RN101 & LSTM & - & - & 60.98 & 62.99 & 59.21 & 48.17 & 52.32 & 42.11  \\
      CMPC~\cite{huang2020referring}     & RN101 & LSTM & - & - & 61.36 & 64.53 & 59.64 & 49.56 & 53.44 & 43.23  \\
      LSCM~\cite{hui2020linguistic} & RN101 & LSTM & - & - & 61.47 & 64.99 & 59.55 & 49.34 & 53.12 & 43.50  \\
      CMPC+~\cite{liu2021cross} &  RN101  & LSTM & - & - & 62.47 & 65.08 & 60.82 & 50.25 & 54.04 & 43.47    \\
      MCN~\cite{luo2020multi} & DN53 & Bi-GRU & 49.22 & 49.40 & 62.44 & 64.20 & 59.71 & 50.62 & 54.99 & 44.69  \\
      BUSNet~\cite{yang2021bottom} & RN101 & Self-Attn & - & - & 63.27 & 66.41 & 61.39 & 51.76 & 56.87 & 44.13  \\
      CGAN~\cite{luo2020cascade}    & DN53 & Bi-GRU & 51.01 & 51.69 & 64.86 & 68.04 & 62.07 & 51.03 & 55.51 & 44.06  \\
      LTS~\cite{jing2021locate}   & DN53 & Bi-GRU & 54.40 & 54.25 & 65.43 & 67.76 & 63.08 & 54.21 & 58.32 & 48.02  \\
      ReSTR~\cite{kim2022restr}  & ViT-B & BERT & - & - & 67.22 & 69.30 & 64.45 & 55.78 & 60.44 & 48.27 \\
      ETRIS~\cite{xu2023bridging} & CLIP-B & CLIP-B & 59.82 & 59.91 & 70.51 & 73.51 & 66.63 & 60.10 & 66.89 & 50.17  \\
      LAVT~\cite{yang2022lavt}  & Swin-B & BERT & 61.24 &  62.09 & 72.73 & 75.82 & 68.79 & 62.14  & 68.38 & 55.10 \\
      SLViT~\cite{OuYang2023SLViTSL} & SegNeXt & BERT & 62.75 & 63.57 & 74.02 & 76.91 & 70.62 & 64.07  & 69.28 & 56.14 \\
      DMMI~\cite{Hu_2023_ICCV} & Swin-B & BERT & 63.46 & 64.19 & \underline{74.13} & \underline{77.13} & \underline{70.16} & 63.98  & 69.73 & \underline{57.03} \\
      DMMI (Reproduced) & Swin-B & BERT & - & - & 73.79 & 75.67 & 69.96 & 63.85  & 69.65 & 55.71 \\
      \midrule
      RISCLIP-B & CLIP-B & CLIP-B & \underline{64.10} & \underline{65.09} & 73.57 & 76.46 & 69.76 & \underline{65.53} & \underline{70.61} & 55.49 \\
      RISCLIP-L  & CLIP-L & CLIP-L & \textbf{67.96} & \textbf{68.71} & \textbf{76.92} & \textbf{80.99} & \textbf{73.04} & \textbf{71.24} & \textbf{76.99} & \textbf{61.56}  \\
      \cmidrule[1.0pt]{1-11}
      \multicolumn{11}{l}{\textbf{mIoU}} \\
      \cmidrule[1.0pt]{1-11}
      CRIS~\cite{wang2022cris}  & CLIP-L* & CLIP-L* & 59.87 & 60.36 & 70.47 & 73.18 & 66.10 & 62.27 & 68.06 & 53.68  \\ 
      SeqTR~\cite{zhu2022seqtr} & DN53 & Bi-GRU & 64.69 & 65.74 & 71.70 & 73.31 & 69.82 & 63.04 & 66.73 & 58.97  \\
      RefTR~\cite{li2021referring} & RN101 & BERT & 66.63 & 67.39 & 74.34 &76.77 & 70.87 & 66.75 & 70.58 & 59.40  \\
      LAVT~\cite{yang2022lavt}  & Swin-B & BERT & {63.34} & {63.62} & {74.46} & {76.89} & {70.94} & {65.81} & {70.97} & {59.23}  \\
      VLT~\cite{Ding_2023} & Swin-B & Bi-GRU & 63.49 & 66.22 & 72.96 & 75.96 & 69.60 & 63.53 & 68.43 & 56.92 \\
      DMMI~\cite{Hu_2023_ICCV}  & Swin-B & BERT & 66.48 &  67.07 & - & - & - & - & - & - \\
      DMMI (Reproduced) & Swin-B & BERT & - & - & 75.26 & 76.96 & 72.05 & 67.51 & 72.1 & 60.38 \\
      \midrule
      RISCLIP-B & CLIP-B & CLIP-B & \underline{67.61} & \underline{67.95} & \underline{75.68} & \underline{78.01} & \underline{72.46} & \underline{69.16} & \underline{73.53} & \underline{60.68}  \\
      RISCLIP-L  & CLIP-L & CLIP-L  & \textbf{71.82} & \textbf{71.65} & \textbf{78.87} & \textbf{81.46} & \textbf{75.41} & \textbf{74.38} & \textbf{78.77} & \textbf{66.84}  \\
      \cmidrule[1.0pt]{1-11}
   \end{tabular}}}
   \vspace{+3mm}
   \label{tab:main_performance}
\end{table*}
\begin{table*}[t!]
  \caption{Comparison with PolyFormer~\cite{liu2023polyformer} in mIoU. Both RISCLIP and PolyFormer are trained on the combined RefCOCO dataset~\cite{yu2016modeling, mao2016generation, nagaraja2016modeling}.
  }
   \centering
   \resizebox{0.95\textwidth}{!}{{\begin{tabular}{lllcccccccc}
      \toprule[1pt]
      \multirow{2}*[-0.5ex]{\textbf{\normalsize{Method}}} & \multirow{2}*[0.5ex]{\textbf{Image}} &
      \multirow{2}*[0.5ex]{\textbf{Text}} & 
      \multicolumn{2}{c}{\textbf{RefCOCOg}} &  \multicolumn{3}{c}{\textbf{RefCOCO}} & \multicolumn{3}{c}{\textbf{RefCOCO+}} \\
      \cmidrule[1.0pt]( r){4-5}
      \cmidrule[1.0pt]( lr){6-8}
      \cmidrule[1.0pt]( lr){9-11}
      & \multirow{2}*[1.5ex]{\textbf{Encoder}} & \multirow{2}*[1.5ex]{\textbf{Encoder}} & Val & Test & Val & Test A & Test B & Val & Test A & Test B  \\
      \cmidrule[1.0pt]{1-11}
      PolyFormer-B~\cite{liu2023polyformer}  & Swin-B & BERT & 69.36 & \textbf{69.88} & 75.96 & 77.09 & \textbf{73.22} & \textbf{70.65} & \textbf{74.51} & \textbf{64.64}\\
      RISCLIP-B  & CLIP-B & CLIP-B & \textbf{69.61} & 69.56 & \textbf{76.01} & \textbf{78.63} & 71.94 & 69.67 & 74.30 & 61.37  \\
      \cmidrule[1.0pt]{1-11}
      PolyFormer-L~\cite{liu2023polyformer}  & Swin-L & BERT & 71.15 & 71.17 & 76.94 & 78.49 & 74.83 & 72.15 & 75.71 & 66.73 \\
      RISCLIP-L & CLIP-L & CLIP-L & \textbf{73.45} & \textbf{74.52} & \textbf{79.53} & \textbf{82.13} & \textbf{75.78} & \textbf{74.88} & \textbf{78.88} & \textbf{68.09}  \\
      \cmidrule[1.0pt]{1-11}
   \end{tabular}}}
   \vspace{+2mm}
   \label{tab:combined}
\end{table*}
\label{sec:experiment_datasets}
\noindent\textbf{Datasets.}
We evaluate RISCLIP on three major RIS datasets: RefCOCO~\cite{yu2016modeling}, RefCOCO+~\cite{yu2016modeling}, and RefCOCOg~\cite{mao2016generation}. The RefCOCO family originates from the same MSCOCO~\cite{Mscoco} dataset and thus shares images but possesses different texts. RefCOCO~\cite{yu2016modeling} and RefCOCO+~\cite{yu2016modeling} texts are relatively concise, consisting of 3.6 words and 1.6 nouns on average. RefCOCO+~\cite{yu2016modeling} differs from RefCOCO~\cite{yu2016modeling} in that the texts do not include absolute positional information, such as first, second, left, and right, is thus more difficult. Lastly, RefCOCOg~\cite{mao2016generation}  comprises of longer, more complex texts (8.4 words and 2.8 nouns per text) and is thus the most challenging. We evaluate on the conventionally used UMD split~\cite{nagaraja2016modeling}.

\noindent\textbf{Evaluation metrics.}
We employ two metrics widely used in RIS: the overall intersection-over-union (oIoU) and the mean intersection-over-union (mIoU). The oIoU is the sum of all intersections over the sum of all unions, while the mIoU is the average of intersection over unions. 
The mIoU is a fairer metric than the oIoU, which is biased towards large objects~\cite{yang2022lavt}. Hence, we report both oIoU and mIoUs but adopt mIoUs when comparing with previous methods. 

\subsection{Model settings}
\label{sec:model_settings}
To explore the effect of the CLIP backbone size, we experiment with two backbones trained with ViT-B and ViT-L~\cite{vaswani2017attention} and dub our framework RISCLIP-B and -L, respectively. In RISCLIP-B, we use the 12-layer ViT-B~\cite{vaswani2017attention} with patch size 16$\times$16 as the image encoder and a 12-layer transformer as the text encoder. In RISCLIP-L, we use ViT-L~\cite{vaswani2017attention} with patch size 14$\times$14 and the same 12-layer text transformer as in RISCLIP-B. For both RISCLIP-B and -L, we attach Adapters in all layers of both encoders, six CFE, and six SKE modules. Other hyperparameters are detailed in Appendix~\ref{sec:training_details}. \par

\subsection{Comparison with state of the arts}
We compare RISCLIP with previous methods on the three aforementioned datasets in Table~\ref{tab:main_performance}. On RefCOCOg-UMD~\cite{nagaraja2016modeling}, RISCLIP-B achieves superior performance compared to DMMI~\cite{Hu_2023_ICCV}, with an average improvement of 1.01 mIoU points across both the validation and test sets. Analogously, RISCLIP-B surpasses DMMI by 0.63 and 1.13 mIoU points on RefCOCO~\cite{yu2016modeling} and RefCOCO+~\cite{yu2016modeling}, respectively. Furthermore, RISCLIP-L, which adopts a larger image encoder, advances the frontier set by RISCLIP-B by an average of 3.96, 3.20, and 5.54 mIoU points, respectively. Such performance improvement across all datasets demonstrates the competency of RISCLIP.

We compare RISCLIP to previous work that leverage CLIP: CRIS~\cite{wang2022cris} and ETRIS~\cite{xu2023bridging}. Different from RISCLIP-L which uses ViT-L~\cite{vaswani2017attention} as the image encoder, CRIS uses ResNet-101~\cite{He_2016_CVPR}. RISCLIP-L surpasses CRIS by an average of 11.62, 8.66, and 11.99 mIoU points on the three datasets, respectively. RISCLIP-B surpasses ETRIS by an average of 4.73, 3.05, and 4.82 oIoU points, respectively. Such performance difference shows that we utilize CLIP effectively. 

Also, we compare RISCLIP to PolyFormer~\cite{liu2023polyformer} in a separate Table~\ref{tab:combined}, since PolyFormer was trained on the combined RefCOCO family while the others were trained on each dataset separately. We also train RISCLIP on the combined dataset following PolyFormer for fair comparison. RISCLIP-B attains comparable performance to PolyFormer-B, but with bigger backbones, RISCLIP-L outperforms PolyFormer-L by an average of 2.83, 2.39, and 2.43  mIoU points on the three datasets. In summary, RISCLIP achieves a new state of the art. 

\subsection{Ablation studies}
We conduct ablation studies on the test set of RefCOCOg-UMD~\cite{nagaraja2016modeling} to prove the effectiveness of our framework and verify its architectural designs. For expedited experiments, we conduct them with a small image size of 240$\times$240 for 50 epochs. Other hyperparameters are the same as those written in Appendix in~\ref{sec:training_details}. \par

\begin{table}[t!]
\caption{Performance when Adapters, CFE, and SKE modules are successively introduced into frozen CLIP. The last row (`Fine-tuned') denotes the setting where CLIP is fine-tuned along with the introduced modules.}
    \centering
    \resizebox{0.49\textwidth}{!}{
    \begin{tabular}{lcccccc}
    \cmidrule[1.0pt]{1-6}
    \textbf{RISCLIP-B} & Adapter & CFE & SKE & mIoU & oIoU &\\
    \cmidrule[1.0pt]{1-6}
    Frozen & \xmark & \xmark & \xmark & 23.86 & 33.13 &\\
    Frozen & \checkmark & \xmark & \xmark & 48.29 & 50.98 &\\
    Frozen & \checkmark & \checkmark & \xmark & 60.58 & 58.39 &\\
    Frozen & \checkmark & \checkmark & \checkmark & 62.64 & 62.02 &\\
    Fine-tuned & \checkmark & \checkmark & \checkmark & 57.88 & 55.75 &\\
    \cmidrule[1.0pt]{1-6}
    \end{tabular}}
    \label{table:ablation_adaptor}
\end{table}
\noindent \textbf{Module ablation.} 
We validate the effectiveness of Adapters, CFE, and SKE modules by progressively introducing each module to frozen CLIP in Table~\ref{table:ablation_adaptor}. Introducing Adapters boosts performance by an mIoU of 24.43, proving that Adapters effectively adapt CLIP to the segmentation task. Moreover, attaching CFE improves performance by 12.29 mIoU, indicating that transforming unimodal feature extraction into a cross-modal one is beneficial for RIS. Introducing SKE further pushes the performance by 2.06 mIoU, showing that comprehensive interaction within CLIP's image-text shared-embedding space is helpful. In contrast, finetuning CLIP along with the modules performs worse than its frozen CLIP twin, with a mIoU drop of 4.76. Thus, our choice of residually adapting frozen CLIP features with Adapters, CFE, and SKE is a viable approach.

\begin{table}[t!]
  \caption{Performance when MHCA in CFE and SKE modules are replaced with other more complex fusion methods, including state-of-the-art fusion mechanisms~\cite{yang2022lavt, Ding_2023}.}
    \resizebox{0.49\textwidth}{!}{
    \begin{tabular}{lccc}
    \cmidrule[1.0pt]{1-4}
    & Fusion Direction & mIoU & oIoU \\
    \cmidrule[1.0pt]{1-4}
    \multicolumn{4}{l}{a) MHCA replaced with complex attention-based fusion modules}\\
    \cmidrule[1.0pt]{1-4}
    MHCA (Ours) & Bidirectional & 62.64 & 62.02 \\
    MHSA on Concat & Bidirectional & 62.63 & 61.65 \\
    MHCA on Concat & Bidirectional & 62.04 & 60.87\\
    \cmidrule[1.0pt]{1-4}
    \multicolumn{4}{l}{b) MHCA replaced with state-of-the-art fusion modules}\\
    \cmidrule[1.0pt]{1-4}
    PWAM~\cite{yang2022lavt} & Text-to-Image &	60.58 & 59.29 \\
    PWAM~\cite{yang2022lavt} & Bidirectional & 61.01 & 59.39 \\
    SDF~\cite{Ding_2023} & Text-to-Image & 59.8 & 57.81 \\
    SDF~\cite{Ding_2023} & Bidirectional & 60.28 & 58.87 \\
    \cmidrule[1.0pt]{1-4}
    \end{tabular}}
    \vspace{-2mm}
    \label{table:ablation_fusion}
\end{table}
\noindent \textbf{Fusion ablation.} 
We validate our choice of a simple MHCA for cross-modal fusion instead of more complex fusion methods. Specifically, we replace the MHCA in CFE and SKE modules with other attention-based fusion methods such as MHSA on concatenated image and text tokens (MHSA on Concat) and MHCA where the query is one modality's tokens and the key, value are the concatenated image and text tokens (MHCA with Concat). Firstly, `MHSA on Concat' produces a slight performance decrease (0.01 mIoU and 0.37 oIoU), indicating it's a viable option. Yet, the computation increase due to attention between the summed number of image and text tokens makes it less efficient than the simple MHCA. Secondly, `MHCA with Concat' decreases performance by 0.6 mIoU and 1.15 oIoU. In summary, the simple MHCA is an efficient yet effective attention mechanism for our CFE and SKE modules. The results are summarized in section a) of Table~\ref{table:ablation_fusion}.

We also demonstrate the superiority of the simple MHCA over existing state-of-the-art fusion modules like Pixel-Word Attention Module (PWAM) in LAVT~\cite{yang2022lavt} and Spatial-Dynamic Fusion (SDF) in VLT~\cite{Ding_2023} for adapting CLIP to RIS. Since both modules perform unidirectional fusion by conditioning image features on text features, we implement bidirectional versions for fair comparison to our CFE and SKE modules that execute bidirectional fusion. As shown in Table~\ref{table:ablation_fusion}, replacing the simple MHCA with these modules in CFE and SKE all results in performance drops, suggesting that complex fusion modules are not needed to adapt CLIP to RIS.

\begin{table}[t!]
  \caption{Performance when the number of Adapters, CFE, and SKE modules are varied. The original setting of 12 Adapters, six CFE, and six SKE modules is marked with asterisk.}
    \resizebox{0.49\textwidth}{!}{
    \begin{tabular}{llccccc}
    \cmidrule[1.0pt]{1-6}
    & Prec@0.5 & Prec@0.7 & Prec@0.9 & mIoU & oIoU &\\
    \cmidrule[1.0pt]{1-6}
    \multicolumn{6}{l}{a) Adapters attached to $N$ last CLIP encoder layers}\\
    \cmidrule[1.0pt]{1-6}
    $N=3$ & 71.81 & 55.3 & 11.29 & 61.40 & 60.84\\
    $N=6$ & 72.73 &	56.53 &	11.87 &	62.15 &	61.33\\
    $N=9$ & 72.50 &	57.44 &	14.03 &	62.31 &	60.68\\
    $N=12$* & 73.19 & 57.68 & 14.21 & 62.64 & 62.02\\
    \cmidrule[1.0pt]{1-6}
    \multicolumn{6}{l}{b) CFE modules attached to $N$ last CLIP encoder layers}\\
    \cmidrule[1.0pt]{1-6}
    $N=2$ &  72.56 & 57.05 & 14.16 &	62.33 &	61.34\\
    $N=4$ &  72.33 & 57.31 & 13.89 &	62.41 &	61.47\\
    $N=6$* & 73.19 & 57.68 & 14.21 & 62.64 & 62.02\\
    \cmidrule[1.0pt]{1-6}
    \multicolumn{6}{l}{c) SKE modules of $N$ layers attached behind CLIP encoders}\\
    \cmidrule[1.0pt]{1-6}
    $N=2$ & 72.17 &	56.72 &	14.23 &	62.30 &	61.56\\
    $N=4$ & 72.73 &	57.68 &	14.57 &	62.79 &	61.95\\
    $N=6$* & 73.19 & 57.68 & 14.21 & 62.64 & 62.02\\
    \cmidrule[1.0pt]{1-6}
    \end{tabular}}
    \vspace{-3mm}
    \label{table:ablation_hyper}
\end{table}
\noindent \textbf{Architecture ablation.} 
We investigate the effect of our modules by varying their numbers in a baseline model. The results are summarized in Table~\ref{table:ablation_hyper}. Section a) shows that performance improves with the number of Adapters attached to the latter CLIP encoder layers. Such a trend suggests that Adapters can beneficially adapt CLIP features to RIS  at all layers. In section b), performance increases with the number of CFE modules, indicating that using more cross-modal interaction during feature extraction is advantageous. In section c), the performance plateaus from 4 to 6 SKE modules, suggesting that there is a limit to the benefits that interaction within the image-text space can bring.

\begin{figure*}[t!]
    \centering
    \includegraphics[width=0.95\linewidth]{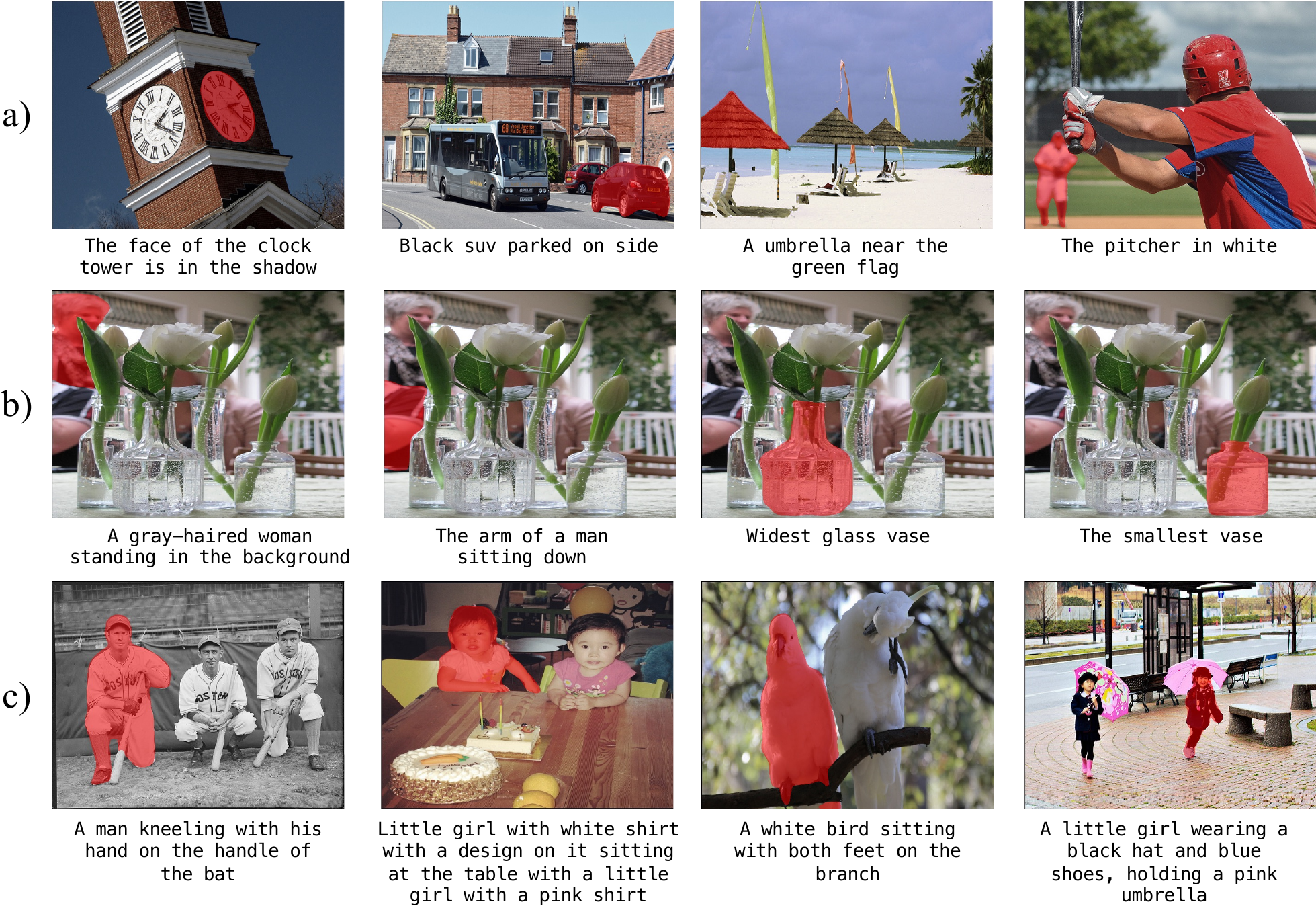}
    \caption{Visualization of RISCLIP-B predictions on RefCOCOg-UMD~\cite{nagaraja2016modeling} test set. Row a) shows RISCLIP's understanding of various instances, row b) RISCLIP's detection of partial, blurry instances and differentiate similar objects, row c) RISCLIP's discernment of the target instance among resembling instances described by lengthy texts.}
    \label{fig:Pretty}
\end{figure*}
\subsection{Visualizations}
We visualize predictions of RISCLIP-B on the RefCOCOg-UMD~\cite{nagaraja2016modeling} test set. Fig.~\ref{fig:Pretty} shows our model's ability to capture a wide variety of instances, detect partially visible or blurry targets, and differentiate the ground truth from resemblances, even with complicated expressions. More visualizations are provided in Appendix~\ref{sec:Visualizations}.

\section{Conclusion}
\label{sec:Conclusion}
RISCLIP effectively extends the image-text alignment of CLIP to RIS, achieving outstanding performance on all major RIS benchmarks. We effectively build upon CLIP's patch-level image-text alignment by introducing cross-modal communication during feature extraction and leveraging the rich cross-modal alignment within CLIP's image-text shared-embedding space to successfully delineate referents described by complicated texts.

\section{Limitations}
\label{sec:Limitations}
We can improve our work by adopting other image-text alignment backbones such as ALIGN~\cite{jia2021scaling} and Florence~\cite{yuan2021florence}. Such extension would allow us to investigate image-text alignment within various cross-modal foundation models and the effectiveness of adapting them to RIS. Also, while RISCLIP achieves state-of-the-art results with impressive margins, there are complex cases where our framework struggles to identify the target instance accurately. We include these cases in Appendix~\ref{sec:failure_cases}.

\section{Broader impacts}
\label{sec:Broader_Impacts}
RIS holds the potential to impact numerous domains that use human-computer interaction, such as autonomous driving and assistant robots. For example, a user could instruct a domestic service robot to ``fetch the blue cup, not the red one'', and the RIS-built-in robot will be able to accurately detect the blue cup and serve his/her owner. Nevertheless, potential ethical concerns, including privacy, model bias, and data processing should be considered. Even the RefCOCO~\cite{yu2016modeling, mao2016generation} dataset includes offensive expressions and provocative images that require removal. In summary, RIS will impact diverse fields adopting human-computer interaction, but ethical issues should be addressed to ensure beneficial development and safe deployment.


\vspace{1mm}
{
\noindent \textbf{Acknowledgement.}
~This work was supported by the NRF grant and 
the IITP grant 
funded by Ministry of Science and ICT, Korea
(NRF-2021R1A2C3012728--35\%, 
 IITP-2019-0-01906--5\%,   
 IITP-2021-0-01343--5\%,   
 IITP-2021-0-02068--20\%,  
 IITP-2022-0-00290--35\%). 
 }

\bibliography{cvlab_kwak}

\appendix

\renewcommand\thefigure{A\arabic{figure}}
\renewcommand{\thetable}{A\arabic{table}}
\setcounter{figure}{0}
\setcounter{table}{0}

\label{sec:supplementary}
\section{Appendices}
\subsection{Training details}
\label{sec:training_details}

\begin{table*}[h!]
  \caption{Hyperparameters for training RISCLIP-B and -L on the separate RefCOCO datasets~\cite{yu2016modeling, mao2016generation, nagaraja2016modeling}. The only difference when training on the combined RefCOCO family~\cite{yu2016modeling, mao2016generation, nagaraja2016modeling} is the batch size, which is increased from 32 to 96 and 56 for RISCLIP-B and -L, respectively. We denote Adam with decoupled weight decay~\cite{loshchilov2019decoupled} as AdamW, rectified linear unit~\cite{agarap2018deep} as ReLU, Brain Floating Point~\cite{burgess2019bfloat16} format as BF16, and single-precision floating-point format as FP32. \vspace{2mm}}
    \centering
    \resizebox{0.73\textwidth}{!}{
    \begin{tabular}{lcc}
    \cmidrule[1.0pt]{1-3}
    \textbf{Hyperparameters} & \textbf{RISCLIP-B} & \textbf{RISCLIP-L}  \\
    \cmidrule[1.0pt]{1-3}
    \multicolumn{3}{l}{\textbf{Backbone}}\\
    \cmidrule[1.0pt]{1-3}
    Pretrained Weight Source & OpenAI & OpenCLIP \\
    Image Encoder Patch Size & 16 & 14 \\
    Image Encoder Transformer Layers & 12 & 24 \\
    Text Encoder Transformer Layers & 12 & 12 \\
    Image Encoder MHA Head Number & 14 & 16 \\
    Text Encoder MHA Head Number & 10 & 12 \\
    $\mathbf{f}_{L}^{v}$ dimension & 896 & 1024 \\
    $\mathbf{f}_{L}^{t}$ dimension & 640 & 768 \\
    $\mathbf{v}$ dimension & 640 & 768  \\
    $\mathbf{t}$ dimension & 640 & 768  \\
    \cmidrule[1.0pt]{1-3}
    \multicolumn{3}{l}{\textbf{Adapters}}\\
    \cmidrule[1.0pt]{1-3}
    Image Encoder Adapter Bottleneck dimension & 449 & 512  \\
    Text Encoder Adapter Bottleneck dimension & 320 & 384  \\
    Non-linear Activation & ReLU & ReLU  \\
    Scaler Initial value & 0.6 & 0.6  \\
    \cmidrule[1.0pt]{1-3}
    \multicolumn{3}{l}{\textbf{Cross-modal Feature Extraction (CFE)}}\\
    \cmidrule[1.0pt]{1-3}
    Module Number & 6 & 6 \\
    $\mathbf{s}_{m-1}^{v}$ & 640 & 768 \\
    $\mathbf{s}_{m-1}^{t}$ & 640 & 768 \\
    MHA Head Number & 10 & 12 \\
    Scaler Initial value & 0.5 & 0.5 \\
    \cmidrule[1.0pt]{1-3}
    \multicolumn{3}{l}{\textbf{Shared-space Knowledge Exploitation (SKE)}}\\
    \cmidrule[1.0pt]{1-3}
    Module Number & 6 & 6 \\
    MHA, MHSA Head Number & 8 & 8 \\
    Scaler Initial value & 0.5 & 0.5 \\
    \cmidrule[1.0pt]{1-3}
    \multicolumn{3}{l}{\textbf{Others}}\\
    \cmidrule[1.0pt]{1-3}
    Image Size & 640 & 560 \\
    Batch Size & 32 & 32 \\
    Epochs & 60 & 60 \\
    Optimizer & AdamW & AdamW \\
    $\beta_{1}$ for AdamW & 0.9 & 0.9 \\
    $\beta_{2}$ for AdamW & 0.999 & 0.999 \\
    Learning Rate Initial Value & 5e-5 & 5e-5 \\
    Weight Decay Strength & 5e-3 & 5e-3 \\
    $\lambda_{\text{dice}}$ & 1.0 & 1.0 \\
    $\lambda_{\text{focal}}$ & 1.75 & 1.75 \\
    $\alpha_{\text{focal}}$ & 0.65 & 0.65 \\
    $\gamma_{\text{focal}}$ & 2.0 & 2.0 \\
    Locator Precision & BF16 & BF16 \\
    Refiner Precision & FP32 & FP32 \\
    \cmidrule[1.0pt]{1-3}
    \vspace{-0.6cm}
    \end{tabular}}
    \label{tab:appendix_hyperparameters}
\end{table*}

\textbf{Training scheme.}
We train both RISCLIP-B and -L for 60 epochs with AdamW~\cite{loshchilovdecoupled} optimizer, using weight decay of 5e-3 and an initial learning rate of 5e-5 with polynomial learning rate decay. Images are resized to 640$\times$640 for RISCLIP-B and 560$\times$560 for RISCLIP-L, such that the visual encoders are both fed 40$\times$40 patch tokens. We apply random affine transformation and random intensity saturation data augmentations following RefTR~\cite{li2021referring}. The ratio between dice~\cite{milletari2016v} and focal loss~\cite{lin2017focal}, $\lambda_{\text{dice}}$ and $\lambda_{\text{focal}}$, is empirically set to 1.0 to 1.75, and alpha and gamma, $\alpha_{\text{focal}}$ and $\gamma_{\text{focal}}$, in the focal loss are set to 0.65 and 2.0. We use batch size of 32 for the models trained on separate RefCOCO datasets~\cite{yu2016modeling, mao2016generation} (reported in Table~\ref{tab:main_performance}), whilst we use bigger batch sizes of 96 for RISCLIP-B and 56 for RISCLIP-L trained on the combined RefCOCO family~\cite{yu2016modeling, mao2016generation} (reported in Table~\ref{tab:combined}) to prevent prolonged training.

\noindent \textbf{Initializations.}
The backbone encoders are initialized from different sources for RISCLIP-B and -L. In RISCLIP-B, the backbone encoders are initialized with the official weights of OpenCLIP~\cite{ilharco_gabriel_2021_5143773} pretrained on LAION-400M~\cite{schuhmann2021laion}. On the other hand, RISCLIP-L's backbone encoders are initialized with the official weights of CLIP~\cite{radford2021learning} pretrained on 400 million image-text pairs collected by OpenAI. We use different sources for the pretrained weights because each source provides a model pretrained with a bigger image size than the other source (\textit{i.e.} OpenCLIP provides a ViT-B backbone pretrained with image size 240$\mathbf{\times}$240 pixels whilst OpenAI provides one with 224$\mathbf{\times}$224 pixels). We empirically find that using a backbone pretrained with a bigger image size provides better segmentation ability. \par
The Adapters adopt different initializations. For the Adapters, we follow \cite{chen2022adaptformer} and initialize the down-projection linear layer with Kaiming Normal~\cite{he2015iccv} and the up-projection layer with zeros. Initializing the up-projection with zeros makes the initial adapter output zero, which is required for stable training~\cite{chen2022adaptformer}. Inspired by this, we also initialize CFE and SKE modules such that the outputs are initially zero. In detail, for CFE modules, we initialize the image-text shared embedding projections in the MHSA as zeros, and, for SKE modules, the value projections in MHA and MHSA as zeros. We experiment with other compositions and find that the adopted initialization provides the best performance, which is slightly better than the others (about 0.6 IoUs).

\noindent \textbf{Additional techniques.}
Furthermore, we observe that incorporating learnable temperatures in the attention modules of the Adapters and introducing learnable channel-wise scalers before residual summation of the Adapter outputs lead to a slight enhancement in performance (up to 0.5 IoU points). All hyperparameters are listed in Table~\ref{tab:appendix_hyperparameters}.

\subsection{Analysis}
In Sections~\ref{sec:failure_cases} and ~\ref{sec:Visualizations}, we analyse RISCLIP-B and RISCLIP-L trained on RefCOCOg-UMD~\cite{nagaraja2016modeling}). We choose RefCOCOg~\cite{mao2016generation} among the three datasets since it possesses longer and more expressive texts, which offer greater insight about the types of texts that RISCLIP understands and struggles with.

\begin{figure*}[t!]
    \centering
    \includegraphics[width=0.9\linewidth]{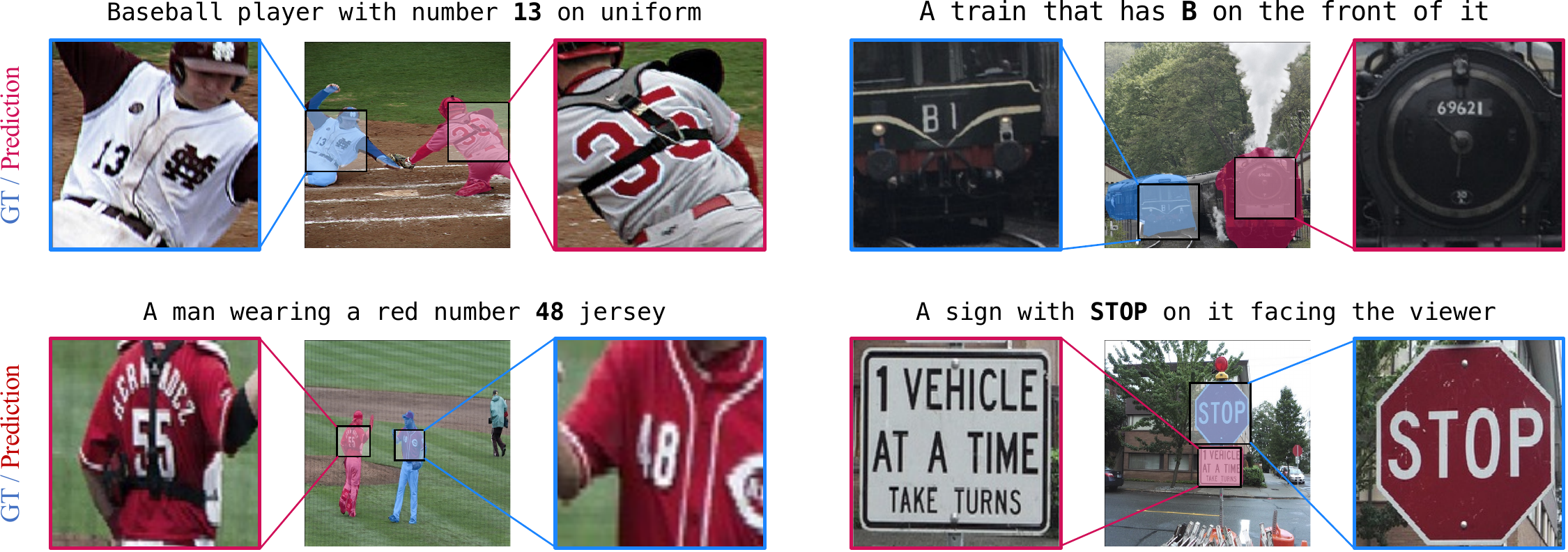}
    \caption{Visualization of RISCLIP-B predictions on RefCOCOg-UMD~\cite{nagaraja2016modeling} test set samples. RISCLIP fails to recognize alphabetic and numeric characters.}
    \label{fig:Failures_Characters}
\end{figure*}

\begin{figure*}[t!]
    \centering
    \includegraphics[width=0.9\linewidth]{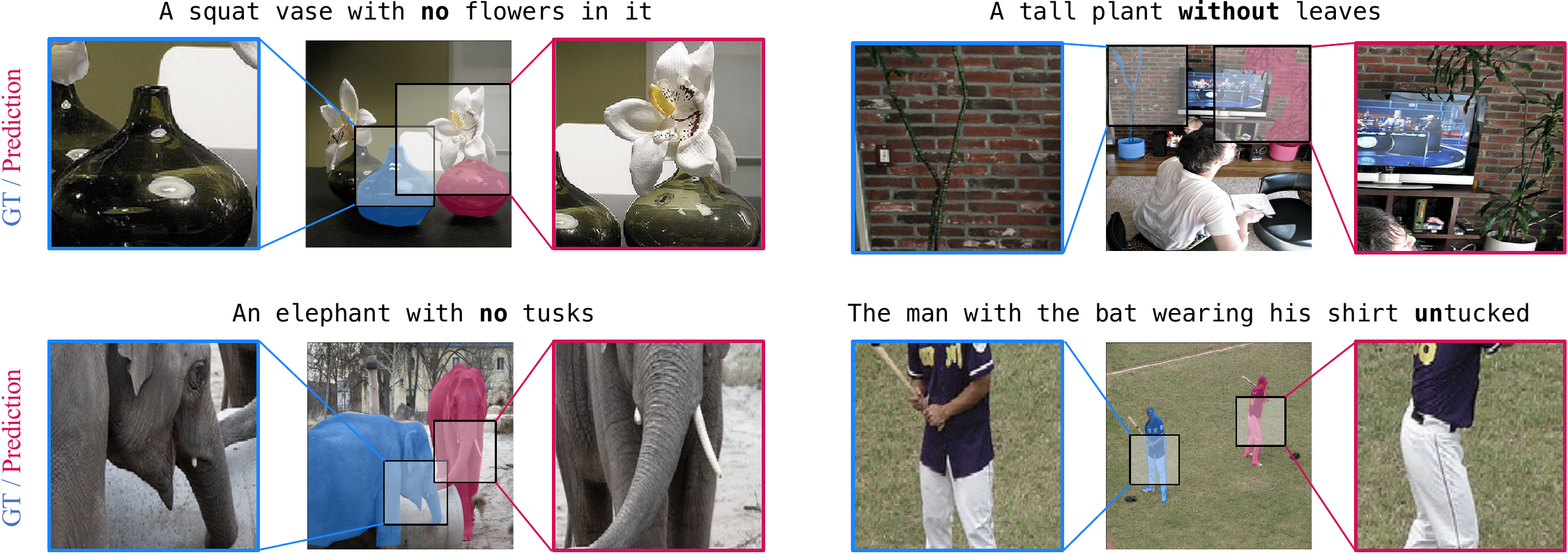}
    \caption{Visualization of RISCLIP-B predictions on RefCOCOg-UMD~\cite{nagaraja2016modeling} test set samples. RISCLIP fails to comprehend texts that describe the target object with the `absence' of some attribute. }
    \label{fig:Failures_Absence}
\end{figure*}

\subsubsection{Failure cases}
\label{sec:failure_cases}
Referring Image Segmentation is a challenging task that involves a various expressions and images. Thus, how to group and categorize the image-text pairs is ambiguous. Nevertheless, we attempt to identify common scenarios where RISCLIP often makes false predictions. Analysing predictions made by RISCLIP-B on the test set, we observe that RISCLIP tends to struggle in two situations: ``Recognition of Characters'' and ``Comprehension of Absence''. We illustrate each case with visualizations, where the ground-truth masks are displayed in blue and predictions made by RISCLIP in pink. \par

\noindent \textbf{Recognition of characters.}
The first case involves the recognition of characters. Figure~\ref{fig:Failures_Characters} shows that RISCLIP fails to detect numbers `13' and `48', the letter `B', and the word `STOP'. \par

\noindent \textbf{Comprehension of absence.}
The second case concerns texts that describe the target instance with the `absence' of some attribute. Figure~\ref{fig:Failures_Absence} shows examples where RISCLIP struggles to comprehend instances described as ``A squat vase with \emph{no} flowers" and ``The man with the bat wearing his shirt \emph{un}tucked". \par

We hypothesize that RISCLIP's relatively poor performance in the two scenarios arises from the limited number of such texts in the dataset. Improving RISCLIP to excel in these cases is another direction for future research.

\subsubsection{Visualizations}
\label{sec:Visualizations}
\begin{figure*}[t!]
    \centering
    \includegraphics[width=0.9\linewidth]{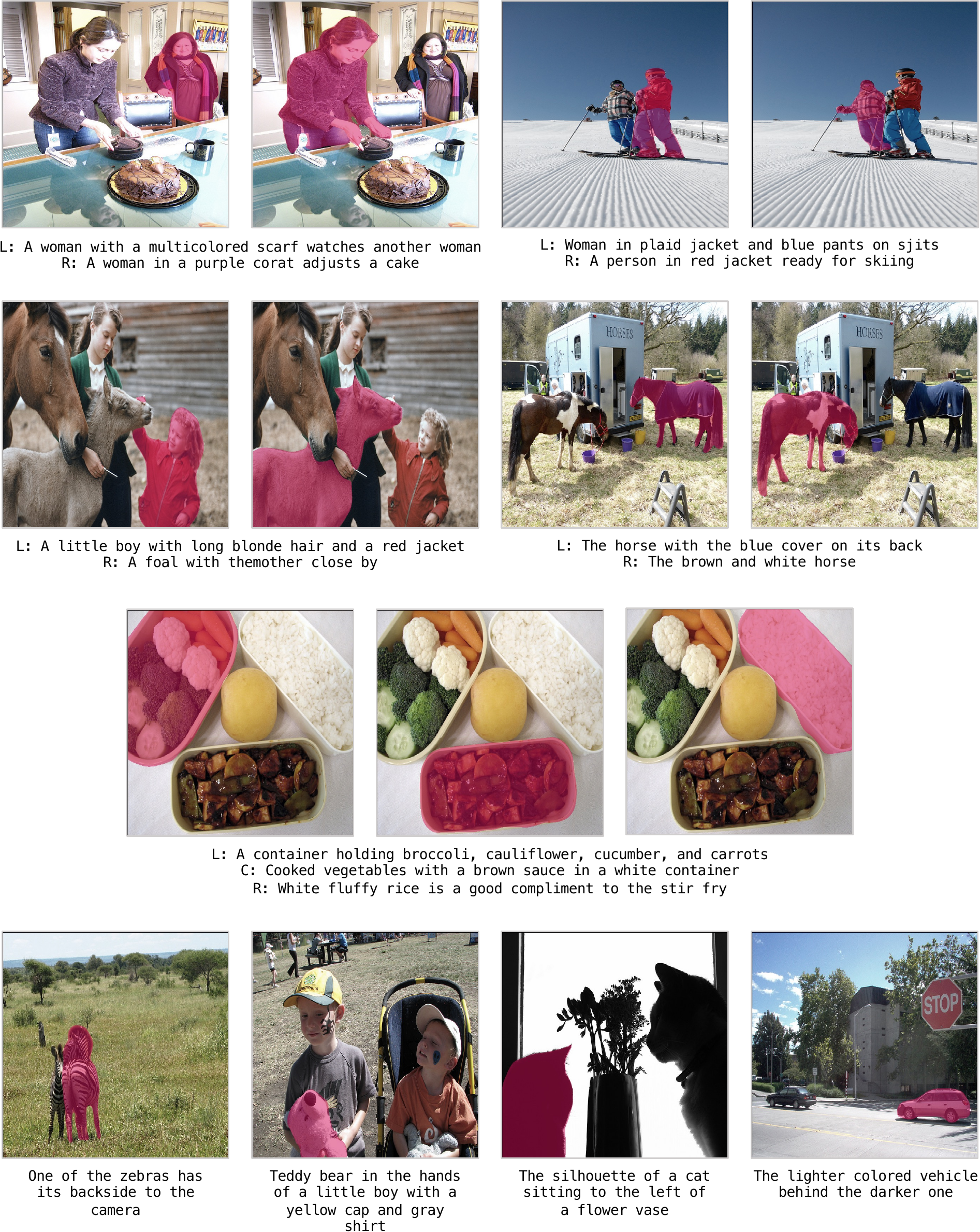}
    \caption{Visualization of RISCLIP-B predictions on RefCOCOg-UMD~\cite{nagaraja2016modeling} test set samples. `L' denotes the text of the left subfigure whilst `R' denotes that of the right. RISCLIP succeeds in locating different target instances within the same image, even when the texts are long and complex. We also present cases where there are similar instances to the target.}
    \vspace{3mm}
    \label{fig:B_successes}
\end{figure*}

\noindent \textbf{RISCLIP-B.}
We provide visualizations of cases where RISCLIP-B successfully segments the target instance on the RefCOCOg-UMD~\cite{nagaraja2016modeling}) test set in Figure~\ref{fig:B_successes}. Even when the texts are lengthy and similar instances exist in the image, RISCLIP-B successfully discerns the referred instance.

\begin{figure*}[t!]
    \centering
    \includegraphics[width=0.9\linewidth]{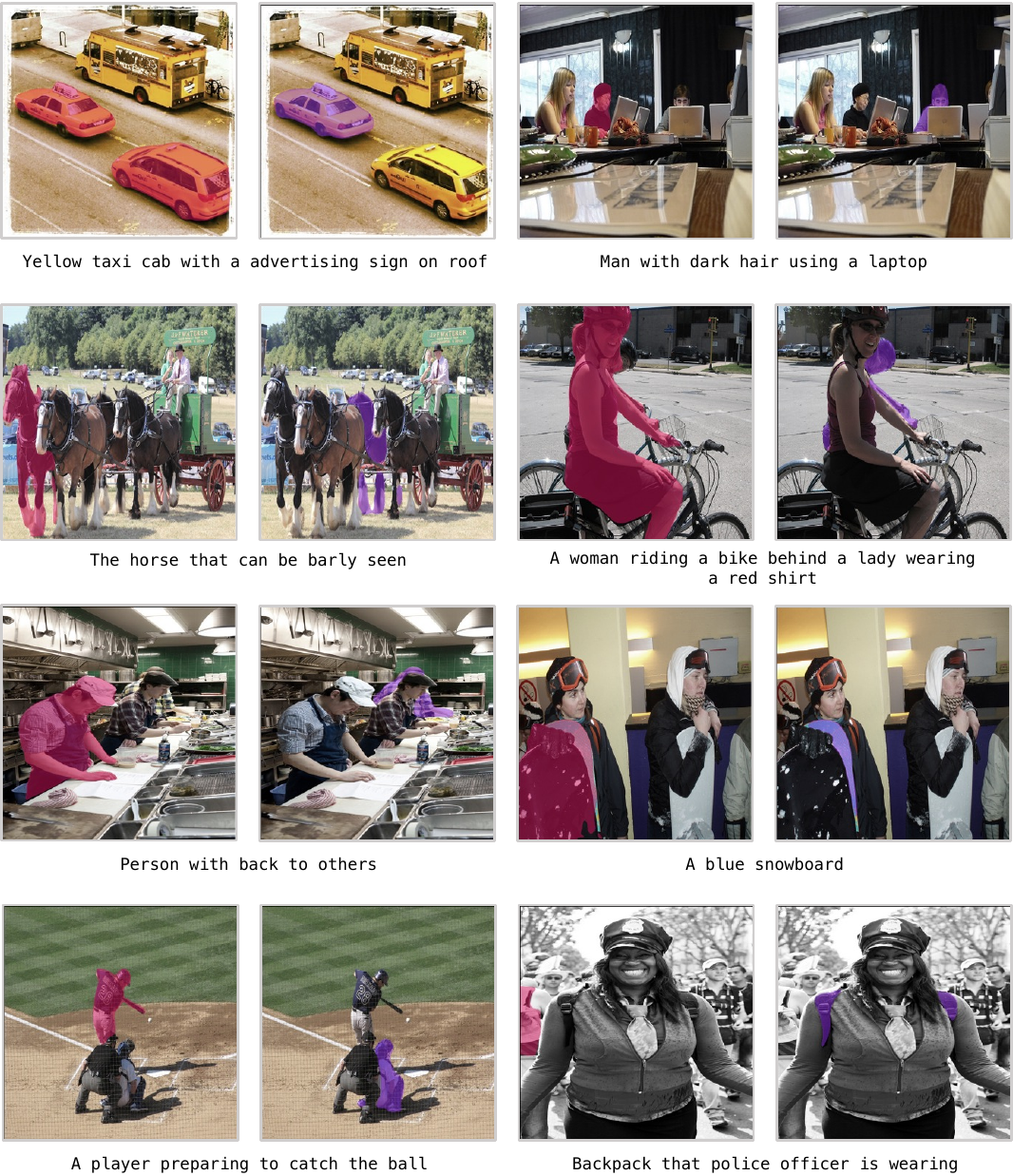}
    \caption{Visualizations of RISCLIP-B (left subfigures in pink) and RISCLIP-L (right subfigures in blue) predictions on RefCOCOg-UMD~\cite{nagaraja2016modeling} test set samples. RISCLIP-L detects instances that have small detecting cues or that are partially visible which are omit by RISCLIP-B.}
    \label{fig:B_vs_L}
\end{figure*}

\noindent \textbf{RISCLIP-L.}
As observed in Table~\ref{tab:main_performance}, RISCLIP-L performs better than RISCLIP-B. 
Thus, we provide visual representations of examples where RISCLIP-L successfully identifies target instances that are overlooked by RISCLIP-B on the RefCOCOg-UMD~\cite{nagaraja2016modeling}) test set in Figure~\ref{fig:B_vs_L}. The segments colored in pink on the left are the predictions made by RISCLIP-B, while the purple segments on the right are those made by RISCLIP-L. \par 
The visualizations suggest that RISCLIP-L possesses an additional capability to detect targets that are only partially visible or require the recognition of subtle visual cues. Such ability can be attributed to the more fine-grained CLIP image encoder of RISCLIP-L: during CLIP~\cite{radford2021learning} pretraining, the CLIP image encoder of RISCLIP-L is trained with image size 336$\times$336 and patch size 14$\times$14 which results in 24$\times$24=576 tokens, whilst that of RISCLIP-B is pretrained with image size 240$\times$240 and patch size 16$\times$16 which amounts to 15$\times$15=225 tokens. Thus, RISCLIP-L possesses are more fine-grained image feature extractor and thereby perceives subtle visual cues better. \par

\end{document}